\def\year{2021}\relax
\documentclass[letterpaper]{article} 
\usepackage{aaai21}  
\usepackage{times}  
\usepackage{helvet} 
\usepackage{courier}  
\usepackage[hyphens]{url}  
\usepackage{graphicx} 
\usepackage{natbib}  
\usepackage{caption} 
\usepackage{tikz}
\usetikzlibrary{positioning, backgrounds, fit, shapes.arrows, shapes.multipart}
\usepackage[shortlabels]{enumitem}

\usepackage{pifont}
\newcommand{\xmark}{\ding{55}}%
\usepackage{subcaption}
\usepackage{amsmath,amssymb,amsfonts}
\usepackage[switch]{lineno}

\title{Fair Representations by Compression}
\author {
        Xavier Gitiaux,
        Huzefa Rangwala \\
}
\affiliations {
George Mason University \\
    xgitiaux@gmu.edu, rangwala@gmu.edu
}

\begin{document}

\maketitle

\begin{abstract}
Organizations that collect and sell data face increasing scrutiny for the discriminatory use of data. We propose a novel unsupervised approach to transform data into a compressed binary representation independent of sensitive attributes. We show that in an information bottleneck framework,  a parsimonious representation should filter out information related to sensitive attributes if they are provided directly to the decoder. Empirical results show that the proposed method, \textbf{FBC}, achieves state-of-the-art accuracy-fairness trade-off. Explicit control of the entropy of the representation bit stream allows the user to move smoothly and simultaneously along both rate-distortion and rate-fairness curves.
\end{abstract}

\section{Introduction}
A growing body of evidence has questioned the fairness of machine learning algorithms across a wide range of  applications, including judicial decisions \cite{ProPublica2016}, face recognition \cite{pmlr-v81-buolamwini18a}, degree completion \cite{gardner2019evaluating} or medical treatment \cite{pfohl2019creating}. Of particular concerns are potential discriminatory uses of data on the basis of racial or ethnic origin, political opinion, religion, or gender. 


Therefore, organizations that collect and sell data are increasingly liable if future downstream uses of the data are biased against protected demographic groups. One of their challenges is to anticipate and control how the data will be processed by downstream users. Unsupervised fair representation learning approaches (\citet{madras2018learning}, \citet{zemel2013learning}, \citet{gitiaux2020learning}, \citet{moyer2018invariant}) offers a flexible fairness solution to this challenge. A typical architecture in fair representation learning includes an encoder that maps the data into a representation and a decoder that reconstructs the data from its representation. The objective of the architecture is to extract from a data $X$ the underlying latent factors $Z$ that correlate with unobserved and potentially diverse task labels, while remaining independent of  sensitive factors $S$. 


 This paper asks whether an encoder that filters out information redundancies could generate fair representations. Intuitively, if sensitive attributes $S$ are direct inputs to the decoder, an encoder that aims for conciseness would not waste code length to encode information related to $S$ in the latent factors $Z$. We show that in an information bottleneck framework \cite{tishby2000information}, this intuition is theoretically founded: constraining the information flowing from the data $X$ to the representation $Z$ forces the encoder to control the dependencies between sensitive attributes $S$ and representations $Z$. \emph{It is sufficient to constraint the mutual information $I(Z, X)$ between $Z$ and $X$ in order to minimize the mutual information $I(Z, S)$ between $Z$ and $S$.} 

Therefore, instead of directly penalizing $I(Z, S)$, we recast fair representation learning as a rate distortion problem that controls explicitly the bit rate $I(Z,X)$ encoded in the latent factors $Z$.  We model the representation $Z$ as a binary bit stream, which allows us to monitor the bit rate more effectively than floating point representations that may maintain redundant bit patterns. 
We estimate the entropy of the code $Z$ with an auxiliary auto-regressive network that predicts each bit in the latent code $Z$ conditional on previous bits in the code. One advantage of the method is that the auxiliary network collaborates with the encoder to minimize the cross-entropy of the code. 



Empirically, we demonstrate that the resulting method, Fairness by Binary Compression (henceforth, \textbf{FBC}) is competitive with state-of-the art methods in fair representation learning. Our contributions are as follows:
\begin{enumerate}
\item We show that controlling for the mutual information $I(Z, X)$ is an effective way to remove dependencies between sensitive attributes and latent factors $Z$, while preserving in $Z$, the information useful for downstream tasks.
\item We find that compressing the data into a binary code as in \textbf{FBC} generates a better accuracy-fairness trade-off than limiting the information channel capacity by adding noise (as in variants of \textbf{$\beta$-VAE}, \cite{DBLP:conf/iclr/HigginsMPBGBML17}). 
\item We show that increasing the value of the coefficient on the bit rate constraint $I(Z, X)$ in our information bottleneck framework allows to move smoothly along both rate-distortion and rate-fairness curves.
\end{enumerate}

\textbf{Related work.}
The machine learning literature increasingly explores how algorithms can adversely impact protected demographic groups (e.g individuals self-identified as Female or African-American) (see \citet{chouldechova2018frontiers} for a review). Research questions revolve around how to define fairness 
(\citet{dwork2012fairness}), how to enforce fairness in standard classification algorithms (e.g. \citet{agarwal2018reductions}, \citet{kim2018fairness}, \citet{kearns2018preventing}) or audit a black box classifier for its fairness (e.g \citet{feldman2015certifying}, \citet{Gitiaux2019mdfaMF}). 

This paper relates to recent efforts towards transforming data into fair and general purpose representations that are not tailored to a pre-specified specific downstream task. Many contributions use a supervised setting where the downstream task label is known while training the encoder-decoder architecture (e.g \citet{madras2018learning}, \citet{edwards2015censoring}, \citet{moyer2018invariant} \citet{song2018learning} or \citet{jaiswal2019discovery}). However, \citet{zemel2013learning}, \citet{gitiaux2020learning} and \citet{locatello2019fairness} argue that in practice, an organization that collects data cannot anticipate what the downstream use of the data will be. In this unsupervised setting, the literature has focused on penalizing approximations of the mutual information between representations and sensitive attributes: maximum mean discrepancy penalty (\citet{gretton2012kernel}) for deterministic (\citet{li2014learning}) or variational (\citet{louizos2015variational}) autoencoders (see Table \ref{tab: recap}); cross-entropy of an adversarial auditor that predicts sensitive attributes from the representations (\citet{madras2018learning}, \citet{edwards2015censoring}, \citet{zhang2018mitigating} or \citet{xu2018fairgan}).

\begin{table*}[]
    \centering
    \begin{tabular}{c|c|c|c}
        Methods & \multicolumn{2}{c|}{Fairness by controlling:} & Examples\\
        & $I(Z, S)$ & $I(Z, X)$ & \\
                \hline
         \textbf{Adversarial} & Minimizing auditor's & \xmark  & \citet{madras2018learning}, \citet{edwards2015censoring}, \\
       & cross-entropy &  & \citet{creager2019flexibly} \\
       \hline
        \textbf{MMD} & Mimizing maximum & \xmark  & \citet{li2014learning}, \citet{louizos2015variational}\\
        & mean discrepancy &  & \\
        \hline
        \textbf{$\beta-$ VAE} & \xmark & Noisy $Z$  & \citet{DBLP:conf/iclr/HigginsMPBGBML17}, This paper\\
        \textbf{FBC} & \xmark & Binary $Z$ & This paper
    \end{tabular}
    \caption{Methods in unsupervised fair representation learning organized by whether the fairness properties of the learned representations is obtained by minimizing the mutual information between sensitive attributes $S$ and representations $Z$; or by minimizing the mutual information between data $X$ and representations $Z$; and whether $Z$ is modelled as a binary bit stream or is convolved with Gaussian noise. 
    }
    \label{tab: recap}
\end{table*}

Our approach contrasts with existing work since it does not control directly for the leakage between sensitive attributes and representations. \textbf{FBC} obtains fair representations only by controlling its bit rate. In a supervised setting, \citet{jaiswal2019discovery} show that nuisance factors can be removed from a representation by over-compressing it. We extend their insights to unsupervised settings and show the superiority of bit stream representations over noisy ones to remove nuisance factors. Our insights could offer an effective alternative to methods that learn representations invariant to nuisance factors (e.g. \cite{JMLR:v19:17-646}, \cite{jaiswal2020invariant}, \cite{NEURIPS2018_03e7ef47}).


Our paper borrows soft-quantization techniques when backpropagating through the model \cite{agustsson2017soft} and hard quantization techniques during the forward pass \cite{Mentzer_2018_CVPR}. 
We find that in our fair representation setting, explicit control of the bit rate of the representation leads to better accuracy-fairness trade-off than floating point counterpart. We estimate the entropy of the code as in \citet{Mentzer_2018_CVPR} by computing the distribution $P(Z)$ of $Z$ as an auto-regressive product of conditional distributions, and by modeling the auto-regressive structure with a PixelCNN architecture (\citet{oord2016pixel}, \citet{van2016conditional}).

\section{Fair Information Bottleneck}
\label{sec: theory}

Consider a population of individuals represented by features $X\in \mathcal{X}\subset[0, 1]^{d_{x}}$ and sensitive attributes in $S\in \mathcal{S}\subset\{0, 1\}^{d_{s}}$, where $d_{x}$ is the dimension of the feature space and $d_{s}$ is the dimension of the sensitive attributes space. In this paper, we do not restrict ourselves to binary sensitive attributes and we allow $d_{s} > 1$. The objective of fair representation learning is to map the features space $\mathcal{X}$ into a $m-$dimensional representation space $\mathcal{Z}\subset[0, 1]^{m}$, such that (i) $Z$ maximizes the information related to $X$, but (ii) minimizes the information related to sensitive attributes $S$. We can express this as  
\begin{equation}
\label{eq: frl}
    \max_{Z}I(X, \{Z, S\})  -  \gamma I(Z, S)
\end{equation}
where $I(X, S)$ and $I(X, \{Z, S\})$  denote the mutual information between $Z$ and $S$ and between $X$ and $(Z, S)$, respectively; and $\gamma \geq 0$ controls the fairness penalty $I(Z, S)$. 

Existing methods focus on solving directly the problem \eqref{eq: frl} by approximating the mutual information $I(Z, S)$ between $Z$ and $S$ via the cross-entropy of an adversarial auditor that predicts $S$ from $Z$ (\citet{madras2018learning}, \citet{edwards2015censoring}, \citet{gitiaux2020learning}) or via the maximum mean discrepancy between $Z$ and $S$ (\citet{louizos2015variational}). 

In this paper, we instead reduce the fair representation learning program \eqref{eq: frl} to an information bottleneck problem that consists of encoding $X$ into a parsimonious code $Z$, while ensuring that this code $Z$ along with a side channel $S$ allows a good reconstruction of $X$. 
The mutual information between $X$ and $S$ can be written as
\begin{equation}\nonumber
\begin{split}
    I(Z, S) &\overset{(a)}{=} I(Z, \{X, S\}) - I(Z, X|S) \\
    & \overset{(b)}{=}I(Z, X) + I(Z, S|X)  - I(Z, X|S) \\
    & \overset{(c)}{=} I(Z, X)- I(Z, X|S) \\
    & \overset{(d)}{=}I(Z, X) - I(X, \{Z, S\}) + I(X, S).
    \end{split}
\end{equation}
where $(a)$, $(b)$ and $(d)$ use the chain rule for mutual information; and, $(c)$ uses the fact that $Z$ is only encoded from $X$, so $H(Z|X, S)=H(Z|X)$ and $I(Z, S|X)=H(Z|X)-H(Z|X, S)=0$. Since the mutual information between $X$ and $S$ does not depend on the code $Z$, the fair representation learning \eqref{eq: frl} is equivalent to the following fair information bottleneck:
\begin{equation}
\label{eq: ib}
    \max\limits_{Z} (1+\gamma)I(X, \{Z, S\}) - \gamma I(Z, X).
\end{equation}
Intuitively, compressing information about $X$ forces the code $Z$ to avoid information redundancy, particularly redundancy related to the sensitive attribute $S$, since the decoder has direct access to $S$. Note that there is no explicit constraint in \eqref{eq: ib} to impose independence between $Z$ and $S$. 

If the representation $Z$ is obtained by a deterministic function of the data $X$, once $X$ is known, $Z$ is known and $H(Z|X)=0$. Therefore, the mutual information $I(Z, X)$ is equal to the entropy $H(Z)$ of the representation $Z$. Since the entropy of the data $X$ does not depend on the representation $Z$, we can replace $I(X, \{Z, S\})=H(X) - H(X|Z, S)$ by $E_{z,s, x}\log(P(x|z, s)$ in the information bottleneck \eqref{eq: ib} and solve for:
\begin{equation}
\label{eq: fib_rd}
    \min\limits_{Z} E_{x, z, s}[-\log(P(X|Z,S)] + \beta H(Z),
\end{equation}
where $\beta =\gamma / (\gamma+1)$. Therefore, the fair representation problem, in its information bottleneck interpretation, can be recast as a rate-distortion trade-off. A lossy compression of the data into a representation $Z$ forces the independence between sensitive attribute and representation but increases the distortion cost measured by the negative log-likelihood of the reconstructed data $E_{x, z, s}[-\log(P(X|Z,S)]$. The parameter $\beta$ in equation \eqref{eq: fib_rd} controls the competitive objectives of low distortion and fairness-by-compression: the larger $\beta$, the fewer the dependencies between $Z$ and $S$.

\section{Proposed Method}
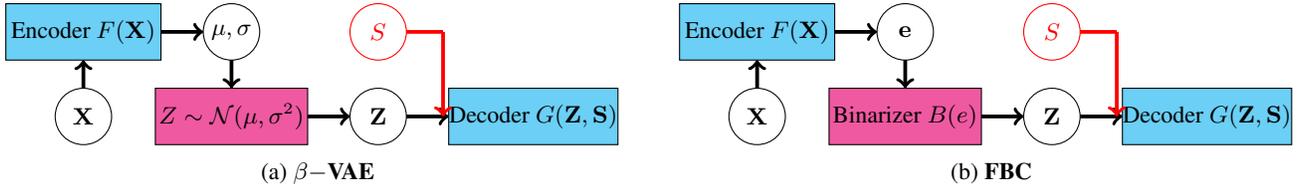
\begin{figure*}
\begin{subfigure}{0.5\textwidth}
\centering
\tikzset{every picture/.style={scale=0.75}}
    \begin{tikzpicture}
\small
\draw (3.125, -1.5) circle (.5cm) node[anchor=center] {$\mathbf{X}$};
\draw[thick, ->, line width=0.5mm] (3.125, -1) -- (3.125, -0.5);
\draw[thick, ->, line width=0.5mm] (4.5, 0) -- (5.25, 0);
\draw[fill=cyan!50] (1.75, -0.5) rectangle (4.5,0.5) node[pos=0.5] {Encoder $F(\mathbf{X})$};
\draw (5.75, 0) circle (.5cm) node[anchor=center] {$\mu, \sigma$};
\draw[thick, ->, line width=0.5mm] (5.75, -0.5) -- (5.75, -1);
\draw[fill=magenta!80] (4.4, -2) rectangle (7.1,-1) node[pos=0.5] {$Z\sim \mathcal{N}(\mu, \sigma^{2})$};
\draw[thick, ->, line width=0.5mm] (7.1, -1.5) -- (7.85, -1.5);
\draw (8.35, -1.5) circle (.5cm) node[anchor=center] {$\mathbf{Z}$};
\draw[thick, ->, line width=0.5mm] (8.85, -1.5) -- (9.6, -1.5);
\draw[fill=cyan!50] (9.6, -2) rectangle (12.6,-1) node[pos=0.5] {Decoder $G(\mathbf{Z}, \mathbf{S})$};
\draw[red] (8.35, 0) circle (.5cm) node[anchor=center] {$S$};
 \draw[red, thick, -, line width=0.5mm] (8.85, 0) -- (9.5, 0);
 \draw[red, thick, ->, line width=0.5mm] (9.5, 0) -- (9.5, -1.5);
\end{tikzpicture}
    \caption{\textbf{$\beta-$VAE}}
    \label{fig: methods1}
    \end{subfigure}
    \begin{subfigure}{0.5\textwidth}
\centering
\tikzset{every picture/.style={scale=0.75}}
    \begin{tikzpicture}
\small
\draw (3.125, -1.5) circle (.5cm) node[anchor=center] {$\mathbf{X}$};
\draw[thick, ->, line width=0.5mm] (3.125, -1) -- (3.125, -0.5);
\draw[thick, ->, line width=0.5mm] (4.5, 0) -- (5.25, 0);
\draw[fill=cyan!50] (1.75, -0.5) rectangle (4.5,0.5) node[pos=0.5] {Encoder $F(\mathbf{X})$};
\draw (5.75, 0) circle (.5cm) node[anchor=center] {$\mathbf{e}$};
\draw[thick, ->, line width=0.5mm] (5.75, -0.5) -- (5.75, -1);
\draw[fill=magenta!80] (4.4, -2) rectangle (7.1,-1) node[pos=0.5] {Binarizer $B(e)$};
\draw[thick, ->, line width=0.5mm] (7.1, -1.5) -- (7.85, -1.5);
\draw (8.35, -1.5) circle (.5cm) node[anchor=center] {$\mathbf{Z}$};
\draw[thick, ->, line width=0.5mm] (8.85, -1.5) -- (9.6, -1.5);
\draw[fill=cyan!50] (9.6, -2) rectangle (12.6,-1) node[pos=0.5] {Decoder $G(\mathbf{Z}, \mathbf{S})$};
\draw[red] (8.35, 0) circle (.5cm) node[anchor=center] {$S$};
 \draw[red, thick, -, line width=0.5mm] (8.85, 0) -- (9.5, 0);
 \draw[red, thick, ->, line width=0.5mm] (9.5, 0) -- (9.5, -1.5);
\end{tikzpicture}
    \caption{\textbf{FBC}}
    \label{fig: methods2}
    \end{subfigure}
    \caption{Unsupervised methods to obtain fair representations $z$ by compression. Variables are: features $X$; sensitive attribute $S$; representation $Z$. $\beta-$VAE generates noisy representations with mean $\mu$ and variance $\sigma^{2}$. FBC generates binary representations. 
    }
    \label{fig: methods}
\end{figure*}

There are two avenues to control for $I(Z, X)$ in the information bottleneck \eqref{eq: ib} (see Figure \ref{fig: methods}): (i) adding noise to $Z$ to control the capacity of the information channel between $X$ and $Z$; or, (ii) storing $Z$ as a bit stream whose entropy is explicitly controlled.

The noisy avenue (i) is a variant of variational autoencoders, so called \textbf{$\beta-$VAE} \cite{DBLP:conf/iclr/HigginsMPBGBML17}, that models the posterior distribution $P(Z|X)$ of $Z$ as Gaussian distributions (see Figure \ref{fig: methods1}). The channel capacity and thus the mutual information between $X$ and $Z$ is constrained by minimizing the Kullback divergence between these posterior distributions and an isotropic Gaussian prior (\citet{braithwaite2018bounded}). In the context of fair representation learning, \cite{louizos2015variational} and \cite{creager2019flexibly} use variants of \textbf{$\beta-$VAE}, but do not focus on how limiting the channel capacity $I(Z, X)$ could lead to fair representations. Instead, they add further constraints on $I(Z, S)$. 

We implement the binary avenue with a method --\textbf{FBC} (see Figure \ref{fig: methods2}) --  that consists of an encoder $F:\mathcal{X}\rightarrow \mathbb{R}^{m}$, a binarizer $B: \mathbb{R}^{m} \rightarrow \{0, 1\}^{m}$ and a decoder $G:\{0, 1\}^{m}\times \mathcal{S}\rightarrow \mathcal{X}$. The encoder $F$ maps each data point $x$ into a latent variable $e=F(x)$. The binarizer $B$ binarizes the latent variable $e$ into a bit stream $z$ of length $m$. The decoder $G$ reconstructs a data point $\hat{x}=G(z, s)$ from the bitstream $z$ and the sensitive attribute $s$. We model encoder and decoder as neural networks whose architecture varies with the type of data at hand.

The binarization layer controls explicitly the bit allowance of the learned representation  and thus forces the encoder to strip redundancies -- including sensitive attributes.
Binarization is a two step process: (i) mapping the latent variable $e$ into $[0,1]^{m}$; (ii) converting real values into 0-1 bit. We achieve the first step by applying a neural network layer with an activation function $\overline{z}=(\tanh(e) + 1) / 2$. We achieve the second step by rounding $\overline{z}$ to the closest integer $0$ or $1$.  One issue with this approach is that the resulting binarizer $B$ is not differentiable with respect to $\overline{z}$. To sidestep the issue, we follow \citet{Mentzer_2018_CVPR} or \citet{theis2017lossy} and rely on soft binarization during backward passes through the neural network. Formally, during a backward pass we replace $z$ by a soft-binary variable $\dot{z}$:
\begin{equation}\nonumber
    \dot{z} = \frac{exp(-\sigma ||\overline{z} -1||_{2}^{2})}{exp(-\sigma ||\overline{z} -1||_{2}^{2}) + exp(-\sigma||\overline{z}||_{2}^{2})},
\end{equation}
where $\sigma$ is an hyperparameter that controls the soft-binarization. During the forward pass, we use the binary variable $z$ instead of its soft-binary counterpart $\dot{z}$ to control the bitrate of the binary representation $Z$ \footnote{In Pytorch, the binarizer returns $(z - \dot{z}).detach() + \dot{z}$.}. 

To estimate the entropy $H(z)$, we factorize the distribution $P(z)$ over $\{0, 1\}^{m}$ by writing $z=(z_{1}, ..., z_{m})$ (\citet{Mentzer_2018_CVPR}) and by computing $P(z)$ as the product of conditional distributions:
\begin{equation}
\label{eq: fac}
    P(z) = \displaystyle\prod_{i=1}^{m}p(z_{i}|z_{i-1}, z_{i-2}, ..., z_{1})\triangleq \displaystyle\prod_{i=1}^{m}p(z_{i}|z_{.<i}),
\end{equation}
where $z_{.<i}=(z_{1}, z_{2}, ..., z_{i-1})$. The order of the bits $z_{1}$, ..., $z_{m}$ is arbitrary, but consistent across all data points. We model $P$ with a neural network $Q$ that predicts the value of each bit $z_{i}$ given the previous values $z_{i-1}, z_{i-2}, ..., z_{1}$. With the factorization \eqref{eq: fac}, the entropy $H(z)$ is given by 
\begin{equation}
\label{eq: ce}
\begin{split}
    H(z) &
    = E_{z}\left[\displaystyle\sum_{i=1}^{m} -\log(Q(z_{i}|z_{.<i}))\right] - KL(P||Q) \\
    & \leq CE(P, Q),
    \end{split}
\end{equation}
where $CE(P, Q)$ is the cross entropy between $P$ and $Q$. Therefore, minimizing the cross-entropy loss of the neural network $Q$ minimizes an upper bound of the entropy of the code $z$. The encoder $F$ and the entropy estimator $Q$ cooperate. The lower the cross-entropy of $Q$ is, the lower is the estimate of the bit rate $H(z)$. Therefore, the encoder has incentives to make the bit stream easy to predict for the neural network $Q$. Designing a powerful predictor for the bit stream $z$ does not necessary complicate the loss landscape, unlike what could happen with adversarial methods (\citet{berard2019closer}). 

Since the prediction of $Q$ for the $i^{th}$ bit depends on the values of the previous bits $z_{i-1}$, ..., $z_{1}$, the factorization of $P(z)$ imposes a causality relation, where the $(i+1)^{th}$, ..., $m^{th}$ bits should not influence the prediction for $z_{i}$. We could enforce this causality constraint by using an iterative method that would first compute $P(z_{2}|z_{1})$, then $P(z_{3}|z_{1}, z_{2})$,..., and lastly, $P(z_{m}|z_{1}, ..., z_{m-1})$. However, it will require $O(m)$ operations that cannot be parallelized. Instead, we follow \citet{Mentzer_2018_CVPR} and enforce the causality constraint by using an architecture for $Q$ similar to PixelCNN (\citet{van2016conditional}, \citet{oord2016pixel}). We model $z$ as a $2D$ $\sqrt{m} \times \sqrt{m}$ matrix and convolve it with one-zero masks, which are equal to one only from their leftmost/top position to the center of the filter. Intuitively, the $i^{th}$ output from this convolution depends only on the bits located to the left and above the bit $z_{i}$. The advantage of using a PixelCNN structure, as noted in \citet{Mentzer_2018_CVPR}, is to enforce the causality constraint and compute $P(z_{i}|z_{.<i})$ for all bits $z_{i}$ in parallel, instead of computing $P(z_{i}|z_{.<i})$ sequentially from $i=1$ to $i=m$.

\section{Experiments}
\begin{figure*}[h]
\centering
\includegraphics[width=\linewidth]{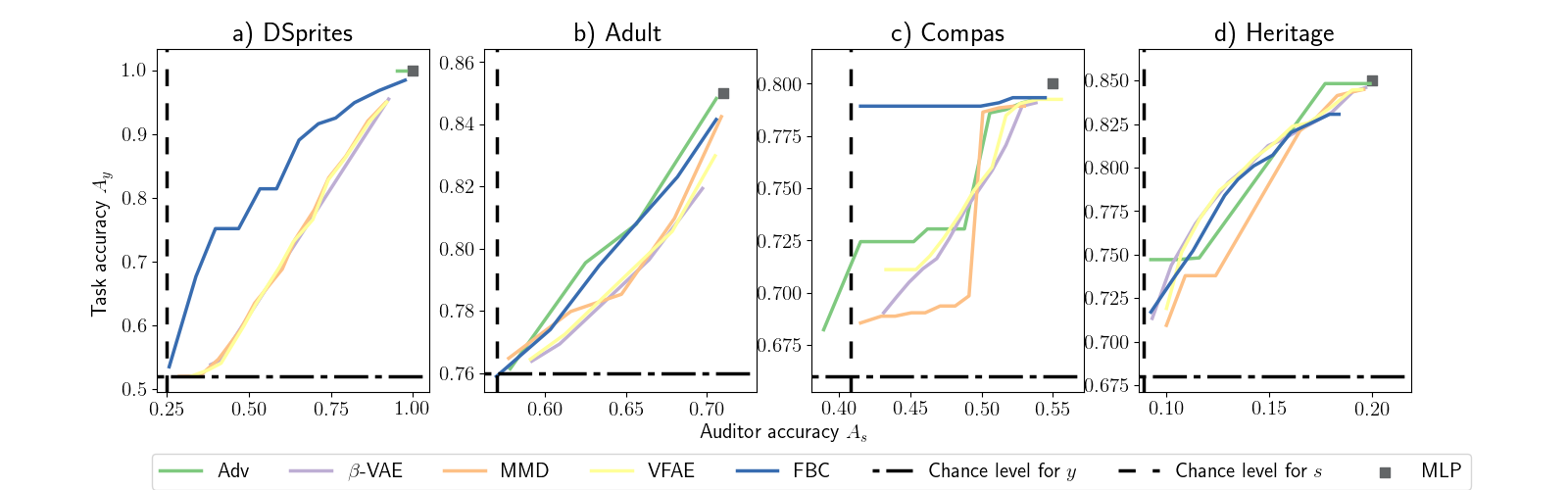}
\caption{Pareto Front for fair representation learning approaches for DSprites and three benchmark datasets. This shows an accuracy-fairness trade-off by comparing the accuracy $A_{s}$ of auditors that predict sensitive attributes $S$ from representations $Z$ to the accuracy of predicting a task label $Y$ from $Z$. The dashed horizontal line represents the chance level of predicting $Y$. The dashed vertical line represents the chance level of predicting $S$. Ranges of $x-$ and $y-$ axes varies across datasets.}
\label{fig: pf}
\end{figure*}
\subsection{Comparative Methods}
The objective of this experimental section is to demonstrate that Fairness by Binary Compression -- \textbf{FBC} -- can achieve state-of-the art performance compared to four benchmarks in fair representations learning: \textbf{$\beta$-VAE}, \textbf{Adv}, \textbf{MMD} and \textbf{VFAE}. 
\begin{enumerate}[(i)]
    \item \textbf{$\beta$-VAE} (\citet{DBLP:conf/iclr/HigginsMPBGBML17}) solves the information bottleneck by variational inference and generates fair representations by adding Gaussian noise which upper-bounds the mutual information between $Z$ and $X$;
    \item \textbf{MMD} (\cite{li2014learning}) uses a deterministic auto-encoder and enforces fairness by minimizing the maximum mean discrepancy (\cite{gretton2012kernel}) between the distribution of latent factors $Z$ conditioned on sensitive attributes $S$;
    \item \textbf{VFAE} (\citet{louizos2015variational}) extends \textbf{$\beta$-VAE} by adding a maximum mean discrepancy penalty; 
    \item \textbf{Adv} (\citet{edwards2015censoring}) uses a deterministic auto-encoder as for \textbf{MMD}, but enforces the fairness constraint by maximizing the cross-entropy of an adversarial auditor that predicts sensitive attributes $S$ from representations $Z$.
\end{enumerate} 
Although \textbf{FBC} shares the deterministic nature of \textbf{Adv} and \textbf{MMD}, it is more closely related to \textbf{$\beta-$VAE}, since \textbf{$\beta-$VAE} obtains fairness without explicit constraint on the mutual information of $I(Z, S)$. The main difference between our approach \textbf{FBC} and \textbf{$\beta-$VAE} is that \textbf{FBC} controls the entropy of a binary coding of the data, while \textbf{$\beta-$VAE} generates noisy representations and approximates the mutual information $I(Z, X)$ with the Kullback divergence between $Q(z|x)$ and a Gaussian prior $P(z)$. Note that the use of a vanilla \textbf{$\beta-$VAE} in a fairness context is novel: only its cousin \textbf{VFAE} with an additional MMD penalty has been proposed as a fair representation method. 

Both \textbf{FBC} and \textbf{$\beta-$VAE} attempt to obtain fairness by controlling $I(Z, X)$. However, \textbf{$\beta-$VAE} assumes further that the prior distribution of the representation is an isotropic Gaussian. \textbf{FBC} does not require such a strong assumption and could still work well even if the data is not generated from a factorized distribution. \textbf{$\beta-$VAE} is meant to compress and factorize. The main result from this paper is that compression is sufficient to learn fair representations and thus, disentanglement might be too restrictive. For problems where factorization could be hard to achieve in an unsupervised setting \cite{locatello2018challenging}, we would expect \textbf{FBC} to outperform \textbf{$\beta-$VAE}. 

\subsection{Experimental Protocol}

The overall experimental procedure consists of:
\begin{enumerate}[(i)]
    \item Training an encoder-decoder architecture $(F, B, G)$ along with an estimator of the code entropy $Q$;
    \item Freezing its parameters;
    \item Training an auditing network $Aud: \mathcal{Z} \rightarrow \mathcal{S}$ that predicts sensitive attributes from $Z$.
    \item Training a task network $T:\mathcal{Z} \rightarrow \mathcal{Y}$ that predicts a task label $Y$ from $Z$. 
\end{enumerate}

The encoder-decoder does not access the task labels during training: our representation learning approach is unsupervised with respect to downstream task labels. Datasets are split into a training set used to trained the encoder-decoder architecture; two test sets, one to train both task and auditing networks on samples not seen by the encoder-decoder; one to evaluate their respective performances. 

\textbf{Pareto fronts.} To compare systematically performances across methods, we rely on Pareto fronts that estimates the maximum information that can be attained by a method for a given level of fairness. We approximate information content as the accuracy $A_{y}$ of the task network $T$ when predicting the downstream label $Y$. The larger $A_{y}$, the more useful is the learned representation for downstream task labels. 

 We measure how much a representation $Z$ leaks information related to sensitive attributes $S$ by the best accuracy $A_{s}$ among a set of auditing classifiers $Aud: \mathcal{Z}\rightarrow \mathcal{S}$ that predict $S$ from $Z$.  The intuition is that if the distributions $p(Z|S=s)$ of $Z$ conditioned on $S$ do not depend on $s$, the accuracy of any classifier predicting $S$ from $Z$ would remain near chance level. In the binary case $\mathcal{S}=\{0, 1\}$, comparing $A_{s}$ to chance level accuracy is a statistical test of independence with good theoretical properties (\citet{lopez2016revisiting}). If the sensitive classes are furthermore balanced ($P(S=0)=P(S=1)$) and the task labels are binary ($\mathcal{Y}=\{0, 1\}$), $A_{s}$  estimates the worst demographic disparity that can be obtained by a downstream task classifier $T$ that uses $Z$ as an input (\citet{gitiaux2020learning}). In the general case $S=\{0, 1\}^{d_{s}}$, the lower $A_{s}$ compared to chance level, the more independent $Z$ and $S$ are. 


\begin{figure*}[htbp]
\centering
\includegraphics[width=\linewidth]{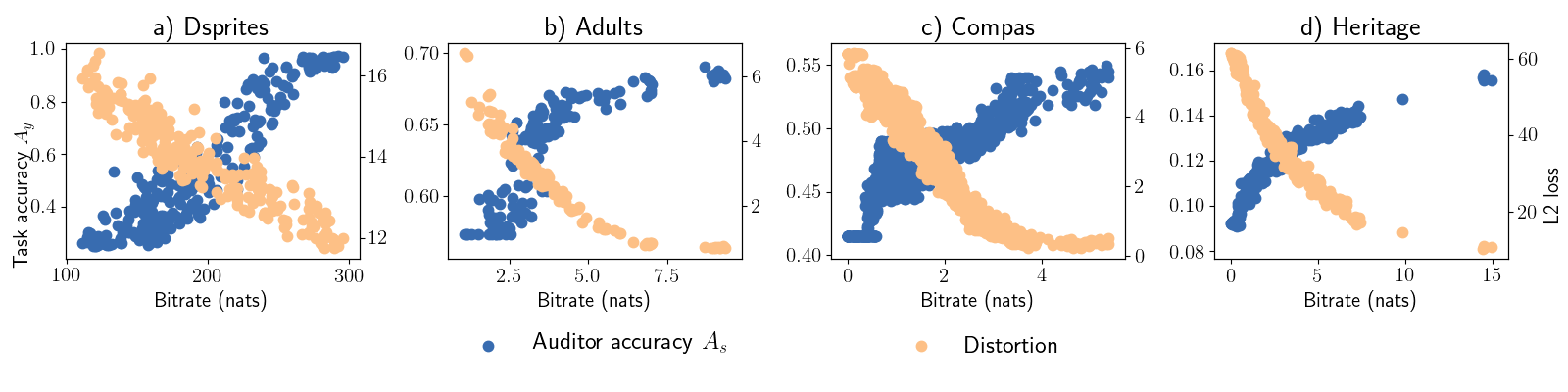}
\caption{Rate distortion/fairness curves. Each dot corresponds to one simulation of FBC. Distortion is measured as the $l2$ loss between reconstructed and observed data.}
\label{fig: rd}
\end{figure*}

\textbf{Rate distortion curves.} To demonstrate further our theoretical insights from section \ref{sec: theory}, we study both rate-distortion and rate-fairness curves of compressing methods \textbf{FBC} and \textbf{$\beta-$VAE}. 

The rate-distortion function $RD(D)$ of an encoder-decoder is measured as the minimum bitrate (in nats) necessary for the distortion $E_{x, z, s}[-\log(p(X|Z,S)]$ to be less than $D$ (\citet{tishby2000information}):
\begin{equation}
    RD(D) = \min I(Z, X) \mbox{ s.t. } E_{x, z, s}[-\log(p(X|Z,S)] \leq D.
\end{equation}

We introduce a new concept, rate-fairness function $RF(\Delta)$, and define it as the maximum bit rate allowed for the accuracy $A_{s}$ of the auditing classifier to remain less than $\Delta$
\begin{equation}
    RF(\Delta) = \max I(Z, X) \mbox{ s.t. } A_{s} \leq \Delta.
\end{equation}

The rate-fairness function captures the maximum information $Z$ can contain while keeping $A_{s}$ under a given threshold. To obtain both rate-distortion and rate-fairness curves for either our binary compression --\textbf{FBC} -- or variational --\textbf{$\beta$-VAE} and \textbf{VFAE} -- approaches , we vary the value of the parameter $\beta$ controlling the rate-distortion trade-off and for each value of $\beta$, we train the model $50$ times with different seeds. For our binary compression method, \textbf{FBC}, the bit rate is approximated by the cross-entropy of the entropy estimator $Q$ in \eqref{eq: ce}; for variational-based methods, the bit rate is approximated by the Kullback divergence between $Q(z|x)$ and a Gaussian prior. In both cases, the approximation is an upper bound to the true bit-rate (in nats) of $Z$. We estimate the distortion generated by the encoder-decoder procedure as the $l_{2}$ loss between reconstructed data $\widehat{X}=G(B(F(X)))$ and observed data $X$.

\textbf{Robustness to Fairness Metrics.}
The fair information bottleneck \eqref{eq: frl} aims at controlling the flow of information between $Z$ and $S$. \cite{mcnamara2017provably} show that minimizing $I(Z, S)$ minimizes an upper bound of the demographic disparity $\Delta(T)$ of a task network $T$ that predicts a binary task label $Y$ from $Z$, where demographic parity $\Delta(T)$ is defined as
\begin{equation}
\label{eq: disp}
   \Delta(T) =  \displaystyle\sum_{s\in \mathcal{S}}|P(T(x)=1|S=s) - P(T(x)=1|S\neq s)|.
\end{equation}

Moreover, the fair information bottleneck \eqref{eq: frl} is solved without a prior knowledge of specific downstream task labels $Y$. Therefore, \eqref{eq: frl} is not designed to control for fairness criteria that rely on labels $Y$ (e.g. equality of odds or opportunites, \cite{NIPS2016_9d268236}) or on a specific classifier (e.g. individual fairness, \cite{dwork2012fairness}), unless downstream task labels are orthogonal to sensitive attributes conditional on features $X$: $Y\perp S|X$. In practice, we explore whether empirically \text{FBC} can generate representations that exhibit for a given task network $T$, low differences in false positive rates $\Delta FPR(T)$ with
\begin{equation}
\label{eq: fpr}
\begin{split}
    \Delta FPR (T) &\triangleq \displaystyle\sum_{s\in \mathcal{S}}|P(T(x)=1|Y=0, S=s)  \\
    & - P(T(x)=1|Y=0, S\neq s)|
\end{split}
\end{equation}

\begin{figure*}[ht]
\centering
\includegraphics[width=\linewidth]{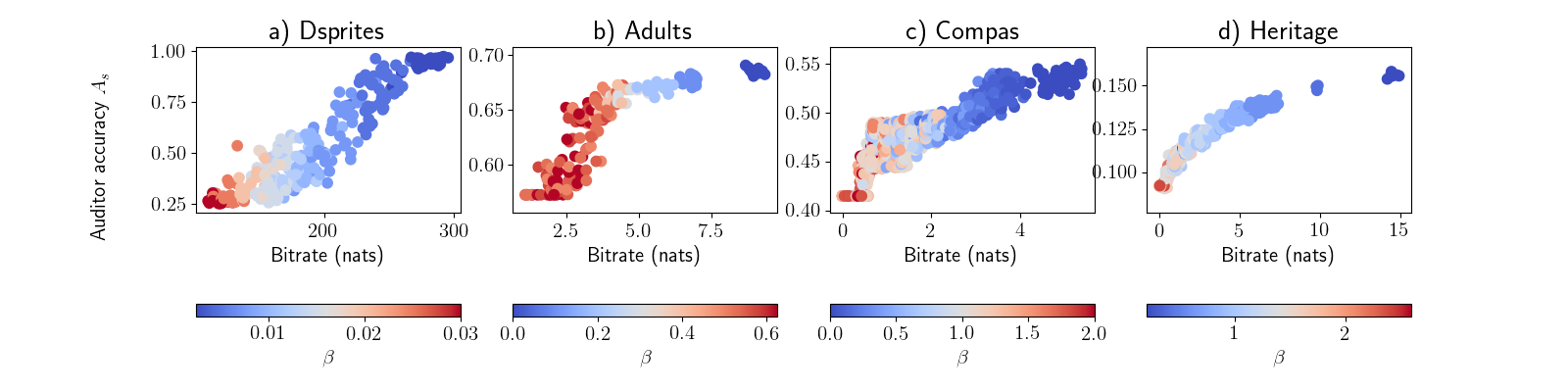}
\caption{Effect of $\beta$. This shows the effect of increasing the coefficient $\beta$ for the code entropy in \eqref{eq: fib_rd} on the bit rate and the auditor's accuracy $A_{s}$ of representations generated by FBC. Changes in $\beta$ allows to move smoothly along the rate-fairness curve.}
\label{fig: beta-effect-fbc}
\end{figure*}
\subsection{Datasets}


    First, we apply our experimental protocol to a synthetic dataset -- DSprites Unfair, \footnote{\url{https://github.com/deepmind/dsprites-dataset/}} -- that contains $64$ by $64$ black and white images of various shapes (heart, square, circle). Images in the DSprites dataset are constructed from six independent factors of variation: color (black or white); shape (square, heart, ellipse), scales (6 values), orientation (40 angles in $[0, 2\pi]$); x- and y- positions (32 values each). We modify the sampling to generate a source of potential unfairness and use as sensitive attribute a variable that encodes the quadrant of the circle the orientation angle belongs to.

Then, we extend our experimental protocol to three benchmark datasets in fair machine learning: \textbf{Adults}, \textbf{Compas} and \textbf{Heritage}. The Adults dataset \footnote{https://archive.ics.uci.edu/ml/datasets/adult}  contains $49K$ individuals and includes information on $10$ features related to professional occupation, education attainment, race, capital gains, hours worked and marital status. Sensitive attributes is made of $10$ categories that intersect gender and race to which individuals self-identify to. 
The downstream task label $Y$ correspond to whether an individual earns more than $50K$ per year.

The Compas data \footnote{https://github.com/propublica/compas-analysis/} contains $7K$ individuals with information related to their criminal history, misdemeanors, gender, age and race. Sensitive attributes intersect self-reported race and gender and result in four categories. The downstream task label $Y$ assesses whether an individual presents a high risk of recidivism.


The Health Heritage dataset \footnote{https://foreverdata.org/1015/index.html} contains $220K$ individuals with $66$ features related to age, clinical diagnoses and procedure, lab results, drug prescriptions and claims payment aggregated over $3$ years. Sensitive attributes are $18$ categories that intersect the gender which individuals self-identify to and their reported age. 
The downstream task label $Y$ relates to whether an individual has a positive Charlson comorbidity Index.

\section{Results and Discussion}
\subsection{Pareto Fronts} Figure \ref{fig: pf} shows the Pareto fronts across five comparative methods for the DSprites and real-world datasets, respectively. Across all dataset, the higher and more leftward the Pareto front, the higher is the task accuracy $A_{y}$ for a given auditor accuracy $A_{s}$ and the better is the accuracy-fairness trade-off. From these Pareto fronts, we can draw three conclusions.

First, on all datasets, controlling for the mutual information between $Z$ and $X$ -- as in \textbf{FBC} and \textbf{$\beta-$VAE} -- is sufficient to reduce the accuracy $A_{s}$ of the auditor $Aud$. This result is consistent with our theoretical observation that minimizing proxies  for the information rate $I(Z, X)$ is sufficient to minimize $I(Z, S)$, provided that a side-channel provides the sensitive attributes $S$ to the decoder.

Second, in the $(A_{s}, A_{y})-$ plan, our method, \textbf{FBC} achieves either similar (Adults, Heritage) or better (DSprites, Compas) accuracy-fairness trade-off than the variational method \textbf{$\beta-$VAE}  that controls $I(Z, X)$ by adding noise to the information channel between $X$ and $Z$. Across all experiments, the Pareto fronts obtained from \textbf{FBC} are at least as upward and leftward as for \textbf{$\beta-$VAE}. This is consistent with our intuition that \textbf{FBC} may outperform \textbf{$\beta-$VAE} in situations where disentanglement of the data into factorized representation is difficult (see \cite{locatello2018challenging} for DSprites).

Third, FBC is a method that appears to be more consistently state-of-the-art in terms of performances compared to existing methods. . \textbf{FBC} offers a better accuracy-fairness trade-off for Compas and DSprites than \textbf{MMD}, \textbf{VFAE} and \textbf{Adv} and is competitive for Adults and Heritage. This is true although  \textbf{Adv}, \textbf{VFAE} and \textbf{MMD} control directly the mutual information between $Z$ and $S$, while \textbf{FBC} controls only $I(Z, X)$. 
The adversarial methods do not manage to generate representations with low $A_{s}$ for the DSprites dataset, possibly because in this higher dimensional problem, the optimization gets stuck in local minima where the adversary has no predictive power, regardless of the encoded representation.


\subsection{Rate-distortion and Rate-fairness}

\begin{figure*}[ht]
\centering
\includegraphics[width=\linewidth]{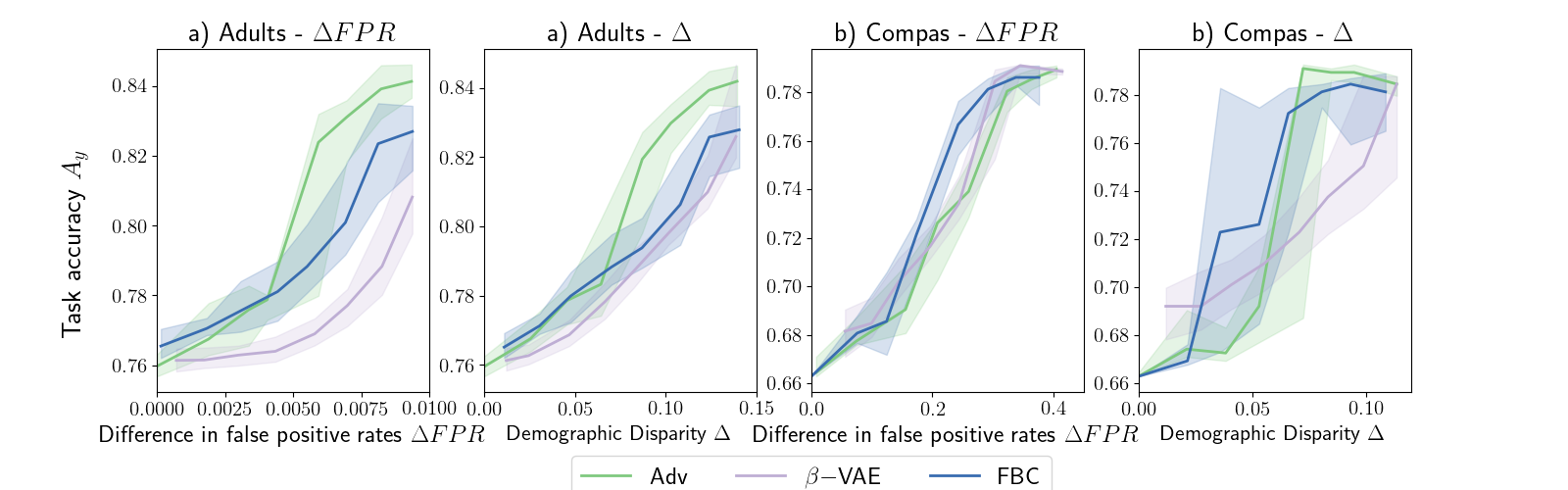}
\caption{Differences in false positive rates and demographic disparity of downstream task networks. This shows pareto fronts for Adults and Compas as in \ref{fig: pf}, but using $\Delta$ (\eqref{eq: disp}) $\Delta FPR$ ( \eqref{eq: fpr}) as a fairness criteria. Shaded areas show the area between the $25-th$ and $75-th$ quantiles of the pareto front.}
\label{fig: other-metrics}
\end{figure*}

\begin{figure}[h]
\centering
\includegraphics[width=\linewidth]{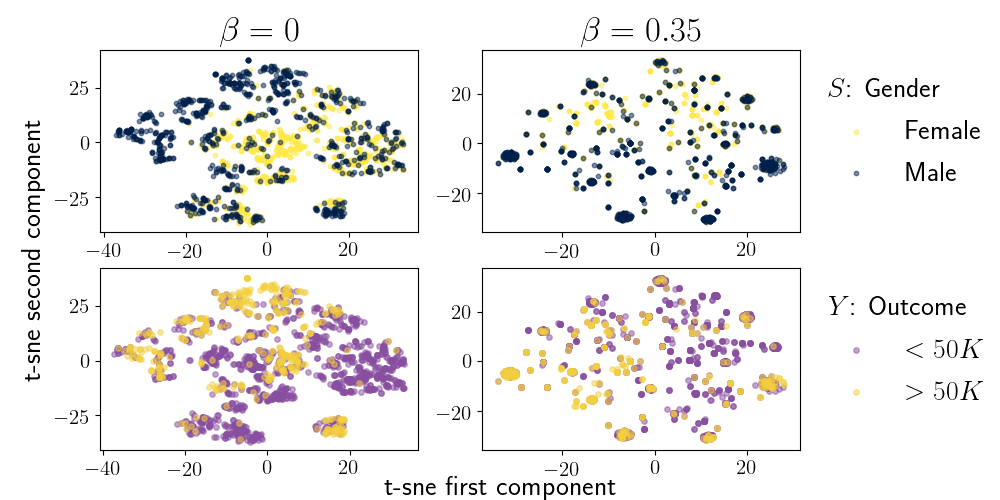}
\caption{Adults – t-SNE visualizations colored with gender ($S$) and income level ($Y$) of the representations obtained by FBC for different values of the parameter $\beta$ controlling the compression rate of FBC.}
\label{fig: tsne}
\end{figure}
Figure \ref{fig: rd} confirms that for \textbf{FBC},  a lower bit rate estimated by the cross entropy $CE(p,q)$ corresponds to a lower accuracy for the auditing classifier $Aud$. Both rate-distortion $(R, D)$ and rate-fairness $(R, \Delta)$ curves show the same monotonic behavior: as distortion moves up along the rate-distortion curves, lack of fairness as measured by $A_{s}$ moves down. However, for real-word datasets, particularly for Adults and Compas, we observe more variance in the auditor accuracy's $A_{s}$ given a representation bit rate. We attribute this higher variance to a smaller sample size -- $617$ for Compas and $3,256$ for Adult on the test set.

Figure \ref{fig: beta-effect-fbc} shows that controlling for the level of compression by increasing the value of $\beta$ in \eqref{eq: fib_rd} allows moving smoothly along the rate-fairness curve. 
This is true whether the mutual information $I(Z, X)$ between data and representation is controlled by the bitstream  entropy as in \textbf{FBC} (Figure \ref{fig: beta-effect-fbc}) or by adding a noisy channel as in \textbf{$\beta-$VAE} (see results in appendix). 
However, binary compression allows a tighter control of the fairness of the representation $Z$ than variational-based methods since in Figure \ref{fig: pf}, for a given auditor's accuracy $A_{s}$,  \textbf{FBC} 
allows the downstream classifier to achieve a higher accuracy $A_{y}$ while predicting $Y$ from $Z$.

\subsection{Other Fairness Metrics}
Figure \ref{fig: other-metrics} extends the pareto fronts of Figure \ref{fig: pf} to additional fairness criteria. It plots the median accuracy obtained by task network $T$ against its differences in false positive rates $\Delta FPR$ and its demographic disparity $\Delta$. 

First, all the methods tested -- \textbf{Adv}, \textbf{$\beta-$VAE} and \textbf{FBC} -- generate an accuracy/fairness trade-off by reducing differences of false positive rates  and demographic disparity at the cost of a lower downstream accuracy. Figure \ref{fig: other-metrics} illustrates a fairness transfer, where general purpose fair representations can offer some guarantees against some fairness criteria that the auto-encoder is not trained to minimize. This transfer is all the more remarkable for differences in false positive rates that rely on downstream task labels $Y$ that were not accessed by the auto-encoder during its training.

Second, for a given value of $\Delta FPR$n or $\Delta$, \textbf{FBC} reaches higher task accuracy $A_{y}$ than \textbf{$\beta-$VAE} and is competitive with \textbf{Adv} for low values of $\Delta FPR$ and $\Delta$.

\subsection{Representation Embeddings}
Figure \ref{fig: tsne} shows the $t-SNE$ visualizations (\citet{maaten2008visualizing}) of the representations generated by \textbf{FBC} for different values of the parameter $\beta$ that controls the rate-distortion trade-off  in \eqref{eq: fib_rd} for the Adults dataset. Without control of the representation bit rate -- $\beta=0$ -- the $t-SNE$ plot show a cluster of Females that are isolated from males and thus, are easily detected by an auditor that predicts $S$ from $Z$. 

With enough compression -- $\beta=0.35$
 -- the representation not only looks more parsimonious, but also does not separate Females from Males as much as without compression ($\beta=0$). In the embeddings space, Females plots are either within  clusters of Males or on the edges of these clusters. Moreover, the $t-SNE$ visualizations separate individuals by income level regardless of the compression level, which confirms that the representations generated by \textbf{FBC} are useful for classification tasks that predict income level from $Z$. $t-SNE$ plots for Compas and Heritage are in the technical appendix.
 
  To quantitatively assess the local homogeneity of the sensitive attribute in the embedding space (Figure \ref{fig: tsne}, top), we compute the average distance of females to their top-10 male neighbors and normalize it by the average distance between all individuals. We find that our homogeneity measure decreases by $30\%$ when compressing the data (from left to right plot). But, a similar measure of homogeneity for outcomes (bottom row) decreases only by $8\%$. This result confirms the visual perception that compression decreases the local homogeneity of sensitive attributes more than the homogeneity of downstream task labels. 

\section{Conclusion}
This paper introduces a new method -- Fairness by Binary Compression (\textbf{FBC}) -- to map data into a latent space, while guaranteeing that the latent variables are independent of sensitive attributes. Our method is motivated by the observation that in an information bottleneck framework, controlling for the mutual information between representation and data is sufficient to remove unwanted factors, provided that these unwanted factors are direct inputs to the decoder. 

Our empirical findings confirm our theoretical intuition: \textbf{FBC} offers a state-of-the-art accuracy-fairness trade-off across four benchmark datasets. Moreover, we observe that encoding the representation into a binary stream allows a tighter control of the fairness-accuracy trade-off than limiting the information channel capacity by adding noise. Our results suggest further research into encoder-decoder whose architecture allows a tighter control of the representation's bit rate and thus, of its fairness.

\section*{Acknowledgments}
This  work  is supported  by  the  National  Science Foundation grant No. 1937950. 
\bibliography{references.bib}

\end{document}


\linenumbers 

\maketitle

\section{Experimental Protocol}
\subsection{Unfair DSprites}
We adapt DSprites to our fairness problem. DSprites (\cite{dsprites17}) contains $64$ by $64$ black and white images of various shapes (heart, square, circle). Images in the DSprites dataset are constructed from six independent factors of variation: color (black or white); shape (square, heart, ellipse), scales (6 values), orientation (40 angles in $[0, 2\pi]$); x- and y- positions (32 values each). The dataset results in $700K$ unique combinations of factor of variations. We modify the sampling to generate a source of potential unfairness. In our baseline experiment, the sensitive attribute is quarternary and encodes which quadrant of the circle the orientation angle belongs to: $[0, \pi/2]$, $[\pi/2, \pi]$, $[\pi, 3/2\pi]$ and $[3/2\pi, 2\pi]$. All factors of variation but shapes are uniformly drawn. When sampling shapes, we assign to each possible combination of attributes a weight proportional to $1 + 10 \left[\left(\frac{i_{orientation}}{40})\right)^{3} + \left(\frac{i_{shape}}{3}\right)^{3}\right]$ , where $i_{shape}\in \{0, 1, 2\}$ and $i_{orientation}=\{0, 1, ..., 39\}$. Since shapes and orientation are correlated, a downstream task predicting shapes could risk to discriminate against some orientation.

\subsection{Architectures.} 

\begin{table}[h]
    \centering
    \begin{tabular}{l|l|l|l|l}
    Dataset & Number of iterations & Learning rate \\
    \hline
    DSprites & 270K &  $10^{-4}$ \\
      \hline
      Adults & 55K & $10^{-3}$ \\
      \hline
       Compas & 22K & $10^{-3}$ \\
       \hline
       Heritage & 55K & $0.5 \times 10^{-4}$ \\
       \hline
\end{tabular}
\caption{Hyperparameter values for FBC.}
\label{tab: hyper}
\end{table}

\paragraph{Encoder-decoders.} For the DSprites dataset, the autoencoder architecture -- taken directly from \cite{creager2019flexibly} -- includes $4$ convolutional layers and $4$ deconvolutional layers and uses ReLU activations. For the three real world datasets, the encoder and decoder are made of fully connected layers with ReLU activations. Table \ref{tab: arch} shows more architectural details for each dataset. For all dataset, the hyperparameter $\sigma$ used for soft-quantization is set to $1$. Other hyperparameter values are in Table \ref{tab: hyper}.

\paragraph{Auditor and task classifiers.} Downstream classifiers and fairness auditors are multi-layer perceptrons with varying width ($64$ to $256$ neurons) and depth ($2$ to $3$ hidden layers).

\paragraph{Entropy estimator}
\begin{figure}[h]
\centering
\tikzset{every picture/.style={scale=0.77}}
    \newcommand*{\GridSize}{4}
\newcommand*{\ColorCells}[1]{
  \foreach \x/\y/\color in {#1} {
    \node [fill=\color, draw=none, thick, minimum size=1cm] 
      at (\x,\GridSize+0.5-\y) {};
    }%
}%

\begin{tikzpicture}
[every node/.style={minimum size=.375cm, outer sep=0pt}]

    \node[fill=gray!50] at (-2.25,1.75) {};
    \node[fill=gray!50] at (-1.75, 1.75) {};
    \node[fill=gray!50] at (-1.25, 1.75) {};
    \node[fill=gray!50] at (-0.75,1.75) {};
    \node[fill=gray!50] at (-0.25,1.75) {};
    \node[fill=gray!50] at (0.25, 1.75) {};
    \node[fill=gray!50] at (0.75, 1.75) {};
    \node[fill=gray!50] at (-2.25,1.25) {};
    \node[fill=gray!50] at (-1.75, 1.25) {};
    \node[fill=gray!50] at (-1.25, 1.25) {};
    \node[fill=gray!50] at (-0.75,1.25) {};
    \node[fill=gray!50] at (-0.25,1.25) {};
    \node[fill=gray!50] at (0.25, 1.25) {};
    \node[fill=gray!50] at (0.75, 1.25) {};
        \node[fill=gray!50] at (-2.25, 0.75) {};
    \node[fill=gray!50] at (-1.75, 0.75) {};
    \node[fill=gray!50] at (-1.25, 0.75) {};
     \node at (-0.75, 0.75) {$z_{i}$};
     \draw[step=0.5cm, color=black] (-3,-2.0) grid (1.5,2.5);
     \draw[pattern=north west lines, pattern color=gray] (-3, 2.0) rectangle (1.5,2.5);
     \draw[pattern=north west lines, pattern color=gray] (-3, -2.0) rectangle (-2.5, 2.5);
     \draw[pattern=north west lines, pattern color=gray] (-2.5, -2.0) rectangle (1.5, -1.5);
      \draw[pattern=north west lines, pattern color=gray] (1.0, -1.5) rectangle (1.5, 2.0);
     \draw[step=0.5cm, color=black, line width=1mm] (-2.5,-1.5) rectangle (1.0, 2.0);
     
     \node[fill=black, text=white] at (3.25,1.75) {1};
     \node[fill=black, text=white] at (3.75,1.75) {1};
     \node[fill=black, text=white] at (4.25,1.75) {1};
     \node[fill=black, text=white] at (3.25,1.25) {1};
     
     \node[text=black] at (3.75,1.25) {0};
      \node[text=black] at (4.25, 1.25) {0};
      \node[text=black] at (3.25, 0.75) {0};
      \node[text=black] at (3.75, 0.75) {0};
      \node[text=black] at (4.25, 0.75) {0};
     
       \draw[step=0.5cm, color=black] (3,0.5) grid (4.5,2.0);
       
    \node[fill=black, text=white] at (3.25, -0.25) {1};
     \node[fill=black, text=white] at (3.75, -0.25) {1};
     \node[fill=black, text=white] at (4.25, -0.25) {1};
     \node[fill=black, text=white] at (3.25, -0.75) {1};
     
     \node[fill=black, text=white] at (3.75, -0.75) {1};
      \node[text=black] at (4.25, -0.75) {0};
      \node[text=black] at (3.25, -1.25) {0};
      \node[text=black] at (3.75, -1.25) {0};
      \node[text=black] at (4.25, -1.25) {0};
      
       \draw[step=0.5cm, color=black] (3,-1.5) grid (4.5,0.0);
\end{tikzpicture}
    \caption{Masks for PixelCNN entropy estimator. Left: binary representations organized into a $\sqrt{m}\times \sqrt{m}-$ 2D structure. Strided cells indicate potential padding with zeros. Top right: filter used in the first layer of our PixelCNN entropy estimator. Bottom right: filter used in the subsequent layers of our PixelCNN entropy estimator.}
    \label{fig: masks}
\end{figure}
The objective of our entropy estimator $Q$ is to predict the $i-th$ bit of the code $Z$ given the values of the previous bits $z_{1}, ..., z_{i-1}$. We organize the representation into a 2D vector, where the order is arbitrary but consistent across data points. Only the bits before the $i^{th}$ bit (shaded in Figure \ref{fig: masks}) are to be used to estimate $P(z_{i}|z_{.<i})$. To do so, we convolve the representations with $c\times c$ filters as in Figure \ref{fig: masks}: top filter for the first convolution; bottom filter for the subsequent convolutions. By repeating these convolutions, we are guaranteed that the resulting feature maps (i) satisfy the causality constraint ($z_{i}$, ..., $z_{m}$ do not affect the $i^{th}$ entry in the feature maps); and (ii) all previous  bits $z_{1}, ..., z_{i-1}$ influence the $i^{th}$ entry of the feature maps. 

In practice, we stack four convolutions with $0-1$ filters as in Figure \ref{fig: masks}. Each convolution layer is followed by batch normalization and ReLU activation. The resulting feature maps are then passed through a traditional convolution layer with a learnable filter and a sigmoid activation function.

\subsection{Pareto Fronts}
To sweep the plan $(A_{y}, A_{s})$ and generate the Pareto fronts, we vary the parameter that controls fairness in each of the competitive methods and for each parameter value, we repeat the experimental protocol $50$ times. We then bin the resulting values of $A_{s}$ and compute the $75\%-$ quantile of $A_{y}$ attained within each bin.

\section{Additional Experimental Results}
\subsection{$\beta-$ VAE}

\begin{figure*}
\centering
\includegraphics[width=\linewidth]{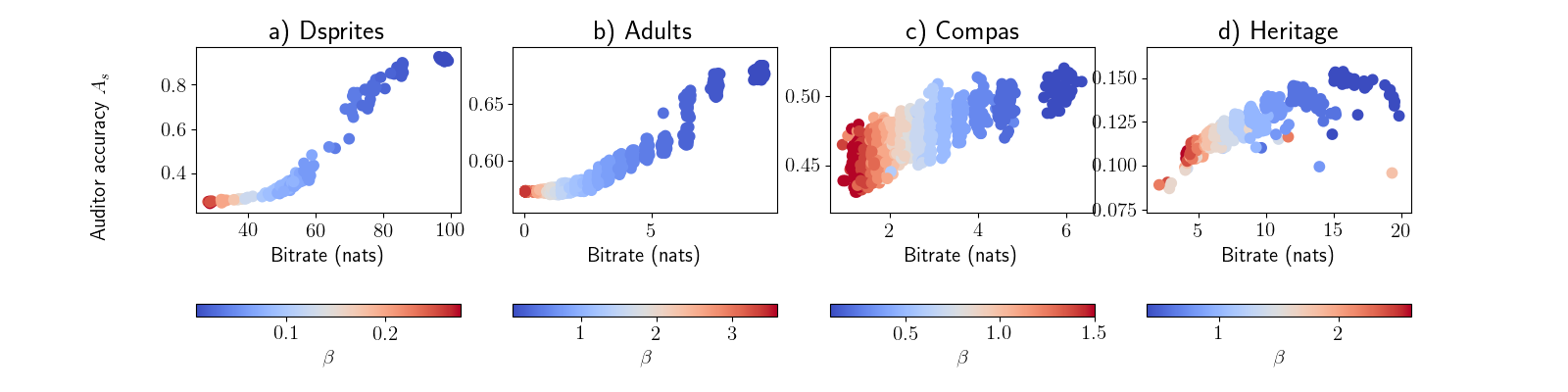}
\caption{Effect of $\beta$ on the fairness of representation generated by \textbf{$\beta-$VAE}. Increasing the coefficient $\beta$ for the Kullback Leibler divergence in \eqref{eq: kl} reduces the bit rate and the auditor's accuracy $A_{s}$ of representations generated by \textbf{$\beta-$VAE}. Changes in $\beta$ allows to move smoothly along the rate-fairness curve.}
\label{fig: beta-effect-vae}
\end{figure*}

The standard \textbf{$\beta-$VAE} (\citet{higgins2016beta}) assumes that the distribution $Q(z|x)$ is Gaussian with mean $\mu(x)$ and standard deviation $\sigma(x)$, and solves for the following minimization problem:

\begin{equation}
\label{eq: kl}
    \min_{q}E_{x, z\sim q(z|x)}[-\log(P(x|z)] + \beta KL(Q(z|x)||P(z)),
\end{equation}
where $P(z)$ is a isotropic Gaussian prior and $KL(Q(z|x)||P(z))$ is the Kullback-Keibler divergence between $Q(z|x)$ and the prior $P(z)$. \citet{higgins2016beta} show that increasing the value of coefficient $\beta$ leads to factorized representation $Z$. We show in our experiments (Figure \ref{fig: beta-effect-vae}) that larger values of $\beta$ also lead to more fair representations, since the Kullback Leibler constraint forces the representations to be more noisy and thus decreases the capacity of the channel between the data and the representation (\citet{braithwaite2018bounded}). This is consistent with our intuition that controlling the mutual information $I(Z, X)$ forces the encoder to filter out redundancies related to sensitive attributes, provided that sensitive attributes are direct inputs to the decoder.

\subsection{$t-SNE$ visualizations}
\begin{figure*}[htpb]
\centering
\includegraphics[width=\linewidth]{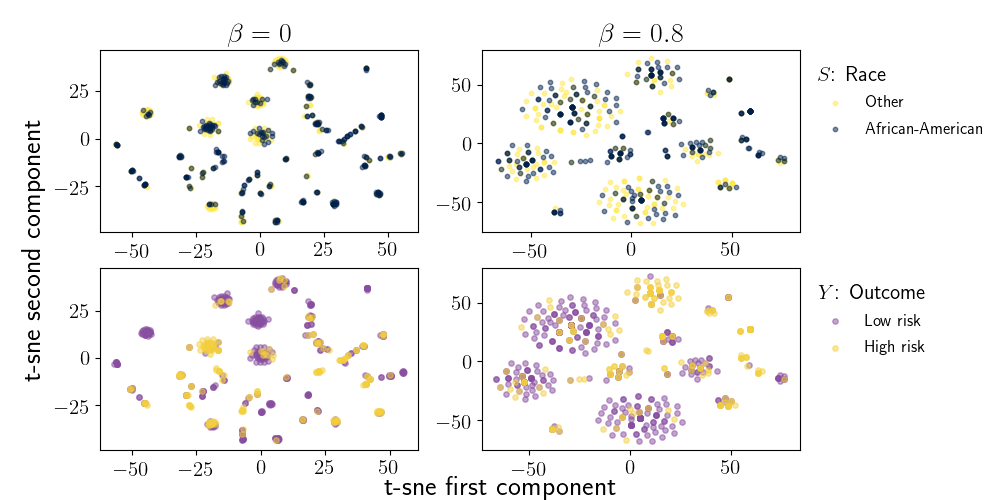}
\caption{Compas – t-SNE visualizations labelled race ($S$, top) and recidivism risk ($Y$, bottom) of the representations obtained by \textbf{FBC} for different values of the parameter $\beta$ controlling the compression rate of \textbf{FBC}.}
\label{fig: tsne-compas}
\end{figure*}

\begin{figure*}[htb]
\centering
\includegraphics[width=\linewidth]{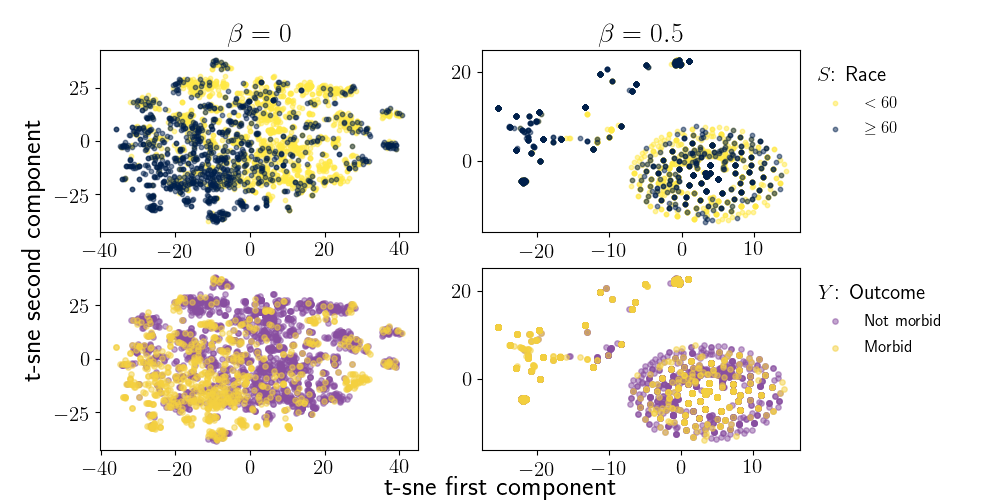}
\caption{Heritage – t-SNE visualizations labelled by age category ($S$, top) and comorbidity index ($Y$, bottom) of the representations obtained by \textbf{FBC} for different values of the parameter $\beta$ controlling the compression rate of \textbf{FBC}.}
\label{fig: tsne-heritage}
\end{figure*}

Figure \ref{fig: tsne-compas} and \ref{fig: tsne-heritage} show $t-SNE$ visualizations for Compas and Heritage, respectively. For both datasets, as $\beta$ increases, representations not only become more concise, but also hide better the protected group which is made of individuals self-identified as African American for Compas (Figure \ref{fig: tsne-compas}, right) or individuals of age $60$ and older (Figure \ref{fig: tsne-heritage}, right) for Heritage.

\begin{table*}[hbpt]
    \centering
    \begin{tabular}{l|l|l|l}
    Dataset & Encoder & Decoder &Activation \\
    & & & \\
    \hline
    DSprites & Conv(1, 32, 4, 2), Conv(32, 32, 4, 2) & Linear(28, 128), Linear(128, 1024) & ReLU \\
     & Conv(32, 64, 4, 2), Conv(64, 64, 4, 2) & ConvT2d(64, 64, 4, 2), ConvT2d(64, 32, 4, 2) &  \\
      & Linear(1024, 128) & ConvT2d(32, 32, 4, 2), ConvT2d(32, 61, 4, 2) &  \\
      \hline
      Adults & Linear(9, 64), Linear(64, 10) & Linear(20, 10), Linear(10, 64), Linear(64,9) & ReLU \\
      \hline
       Compas & Linear(6, 16), Linear(16, 8) & Linear(12, 8), Linear(8, 16), Linear(16,6) & ReLU  \\
       \hline
       Heritage & Linear(65, 128), Linear(128, 24) & Linear(42, 24), Linear(24, 128), Linear(128,65) & ReLU \\
       \hline
\end{tabular}
\caption{\textbf{Architecture details.} $Conv2d(i, o, k, s)$ represents a 2D-convolutional layer with input channels $i$, output channels $o$, kernel size $k$ and stride $s$.  $ConvT2d(i, o, k, s)$ represents a 2D-deconvolutional layer with input channels $i$, output channels $o$, kernel size $k$ and stride $s$. $Linear(i, o)$ represents a fully connected layer with input dimension $i$ and output dimension $o$.}
\label{tab: arch}
\end{table*}

\pagebreak
\bibliography{references.bib}